\documentclass{article}
\usepackage{spconf,amsmath,epsfig}
\usepackage{gensymb}
\usepackage{capt-of}
\usepackage{booktabs}

\renewcommand{\d}[1]{{\mbox{\boldmath$#1$}}}

\title{Sea Level Anomaly Prediction using Recurrent Neural Networks}
\name{Anne Braakmann-Folgmann, Ribana Roscher, Susanne Wenzel, Bernd Uebbing and J\"urgen Kusche
}
\address{\small Institute of Geodesy and Geoinformation, University of Bonn, Nussallee 15, D-53115 Bonn.
\\
\small 
Contact: \{abraakmann / rroscher / susanne.wenzel / bernd.uebbing / kusche\}@uni-bonn.de 
}
\begin{document}
%\ninept
%
\maketitle
\begin{abstract}
Sea level change, one of the most dire impacts of anthropogenic global warming, will affect a large amount of the world's population. However, sea level change is not uniform in time and space, and the skill of conventional prediction methods is limited due to the ocean's internal variabi-lity on timescales from weeks to decades. Here we study the potential of neural network methods which have been used successfully in other applications, but rarely been applied for this task.
We develop a combination of a convolutional neural network (CNN) and a recurrent neural network (RNN) to ana-lyse both the spatial and the temporal evolution of sea level and to suggest an independent, accurate method to predict interannual sea level anomalies (SLA). 
We test our method for the northern and equatorial Pacific Ocean, using gridded altimeter-derived SLA data.
We show that the used network designs outperform a simple regression and that adding a CNN improves the skill significantly. The predictions are stable over several years. 
\end{abstract}

\section{Introduction}
\label{sec:intro}

Modelling and predicting sea level anomalies (SLA) is currently a relevant topic, as sea level responds to global warming directly by thermal expansion of the ocean and indirectly by mass increase through melting ice sheets and glaciers. 
And with 60\% of the world's population living in coastal areas an accurate prediction of SLAs at a high spatial resolution is required. Sea level rise is both spatially and temporally highly variable and therefore especially hard to model. 
Predictions by ocean models rely on incomplete representation of physical processes, limited spatial resolution, uncertain initial conditions, and scenario-based boundary conditions. 

Complementing physical ocean modelling, we develop a new approach using neural networks. Neural networks are a powerful mean to solve tasks such as classification, object detection and speech recognition, where they show promising results reaching accuracies superior to classical and shallow state-of-art machine learning algorithms \cite{c22}. 

So far, they have barely been explored in the context of sea level prediction. 
Makarynskyy et al. \cite{harbor} use a single tide gauge to predict the local sea level in a harbour, and Wenzel and Schr\"oter \cite{c10} use global tide gauge data, to derive average trends and amplitudes for eight different regions. However, they lack a high spatial resolution and the used one-layered fully connected network architectures are simple compared to current networks applied in the computer vision community. 
 
Here, in contrast mostly convolutional neural networks (CNNs) and very deep networks are used, as they reach higher accuracies \cite{c22}. CNNs are particularly suitable for gridded data exploiting the spatial correlations. On the other hand recurrent neural networks (RNNs) and their enhancement Long Short Term Memory networks (LSTM, \cite{c11}) are specifically designed to model time series data. A combination of CNN and RNN has for example been used to predict one frame ahead in video sequences \cite{s18}. With sequence-to-sequence LSTMs \cite{s23} it is possible to predict several time steps ahead. 

Our aim is first to design a sophisticated network composed of a CNN and an RNN to capture both spatial and temporal relations of radar-altimetric SLA fields in the Pacific Ocean and to predict the SLA map of the next month. Then we extend the prediction length to several years.

\section{Data}
\label{sec:pagestyle}

We use time series of SLAs from ESA (European Space \linebreak Agency) CCI (Climate Change Initiative) \cite{c1} as input to our network. SLAs are the difference between actual sea surface height (SSH) and mean sea surface height. We use the Level~4 product of ESACCI, where all altimetry mission measurements have been merged into monthly grids with a spatial resolution of 1/4 degree. The data cover 23 years from \mbox{January} 1993 to December 2015. To validate the predictive skill, we split the data into training- (16 years), validation- (4~years) and test data (3 years). 

The northern and equatorial Pacific make a perfect test region, since parts of it undergo rapid sea level rise while being affected by strong interannual and decadal variability \cite{riet}. Here we examine the region between 110\degree and 250\degree ~longitude and 15\degree S to 60\degree N latitude. 
This region holds 170,240 grid points, which give us almost 33 million values for training of the network and 47 million values overall.

\begin{figure*}[h!]
\centering
	\includegraphics[width=0.91\textwidth]{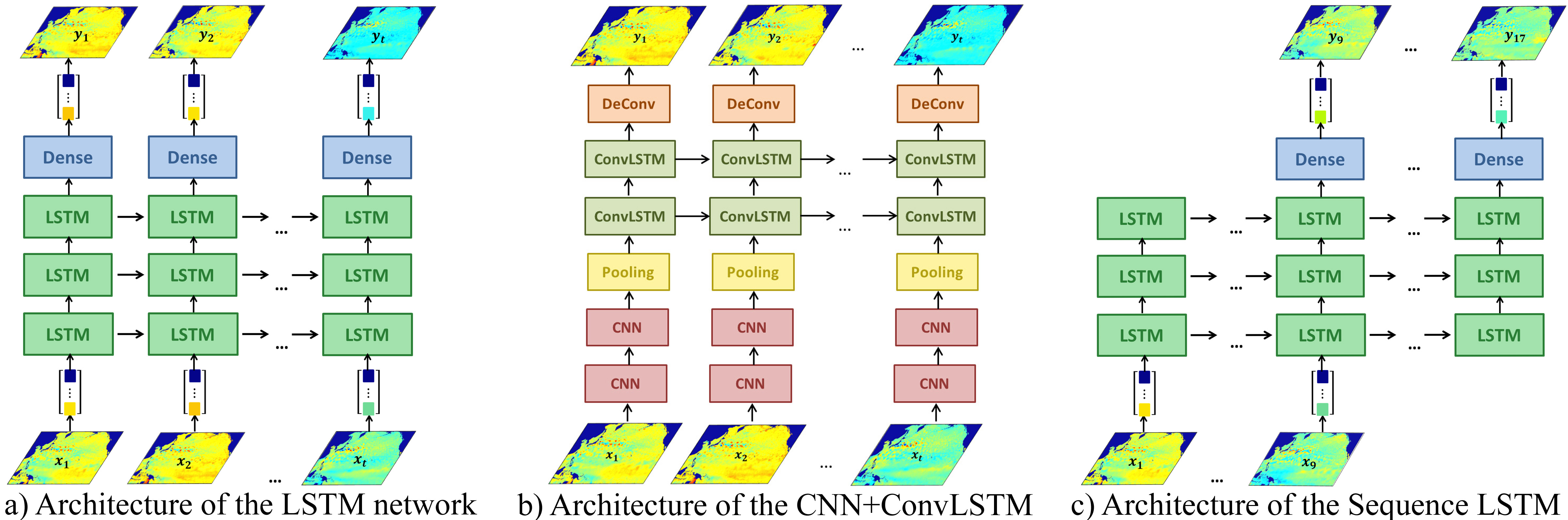}
\vspace{-0.2cm}
\caption{Different network designs. The input \d x is all SLAs at one time step t. In a) and c) they are reshaped to a vector, in b) the grid structure is preserved. The network predicts the vector/grid \d y at the next one month (in a) and b)) or nine months (in c)) 
}
\label{fig:seq}
\end{figure*}

\section{Methodology}
\label{sec:typestyle}

SLAs reflect various complex non-linear interactions in the ocean
and hence contain several deterministic and stochastic modes that are difficult to predict. Yet, SSH maps contain many reoccuring spatial patterns (e.g. eddy fields, gyres, ENSO) that can potentially be learned by neural networks.

Artificial neural networks are a means of machine learning inspired by the human brain to learn higher-order representations  and perform diverse tasks. In contrast to other machine learning techniques, neural networks are able to extract relevant features and their weight in the model. CNNs have shown to be efficient networks which are especially designed to capture spatial dependencies by using gridded input data.

RNNs are neural networks that can deal with time series analysis, taking sequences as in- and output \cite{s23}. In contrast to feedforward networks they incorporate a self-loop, which enables the net to memorise the previous inputs (horizontal information flow in Fig. 1). LSTMs \cite{c11} are a special kind of RNN that are even capable of learning long-term dependencies.
Convolutional LSTMs (ConvLSTM, \cite{c30}) preserve the input's spatial structure by replacing all matrix multiplications within the LSTM with convolutions. 
To learn higher-order representations, both RNNs and CNNs can be stacked on top of each other.

\section{Experimental Setup}

We design different models using the framework Keras \cite{keras} with Tensorflow backend. 
We train them with the Adam optimizer \cite{c12} to minimise the MSE between their prediction and the truth in 150 epochs using decaying learning rates.
\vspace{0.2cm}

{\bf LSTM network:} 
Here the input is a vector with all SLAs of one month. The network consists of three LSTM layers with 60 units each, followed by a fully connected layer that learns how to make a prediction of the next month for each 1/4\degree grid cell from the 60 features (Fig. 1a). This prediction is our output. The network consits of over 51 million parameters to train. We employ a hard sigmoid activation function for the hidden-to-hidden state translations and a tanh activation function in the LSTMs' hidden-to-output transformation, allowing the net to learn non-linear relations. To further improve the generalisation capability we employ LSTM-modified dropout \cite{rnndrop} of 0.8. On an Intel i5-2400 CPU training takes 8~hours. After each prediction of one month we use the true values of the preceding month as input for the next prediction. 

{\bf CNN+ConvLSTM network:} Here the input SLAs are used in their natural grid structure, preserving spatial information. In this network we combine a CNN with a ConvLSTM as shown in Fig.~1b. The CNN consists of two convolutional layers with 32 filters, a kernel size of 3 x 3 and a ReLU activation function each, followed by a pooling layer with kernel size 4 x 4 and stride 4. 
We use the extracted feature maps as input to
two ConvLSTM layers with 40 units each, a kernel size of 3 x 3 and the same activation functions as used in the LSTM. 
The last layer is a deconvolutional (DeConv, \cite{deconv}) layer to regain the spatial extent of our input and to map the 40 feature maps to a single SLA value at each grid cell. This map of SLAs is our output prediction for the next month. Each predicted grid cell has a receptive field of 6\degree. This net has to learn only 229,413 parameters due to pooling. We apply batch normalisation to the input. Dropout is not used. On an Intel i5-2400 CPU training this net takes 18~hours.

{\bf Sequence LSTM network:} 
This network is equal to the LSTM network, but takes sequences of nine months as in- and output (see Fig. 1c). While the former two networks only predict one month in advance, this network is able to predict the following nine months at once. After the prediction of a nine months long sequence, we use the true values of the preceding nine month as input for the next prediction. On the Intel i5-2400 CPU training this net takes 8~hours, too.

{\bf Sequence LSTM-P network:} This network is trained just like the Sequence LSTM. However here we take the predicted nine months as input to predict nine months further.

\section{Results}
\label{sec:majhead}

In this section we assess the networks' performance and compare the predictions made by our networks to a regression approach. We use the sum of a trend, acceleration, annual and semi-annual sine and cosine as regression model. The regression parameters are estimated for each grid cell individually.

Figure~2 shows the measured time series of SLAs at one point in the North Pacific Gyre.
The position is marked with a cross in Fig. 3. 
We visualize the prediction during training (blue), validation phase (green) and the actual test phase (red), compared to the known true data (black). 

\vspace{0.6cm}

\hspace{-0.65cm}
\begin{minipage}[t!]{0.45\textwidth}
\begin{center}
%\vspace{-5cm}
	\includegraphics[width=0.85\textwidth]{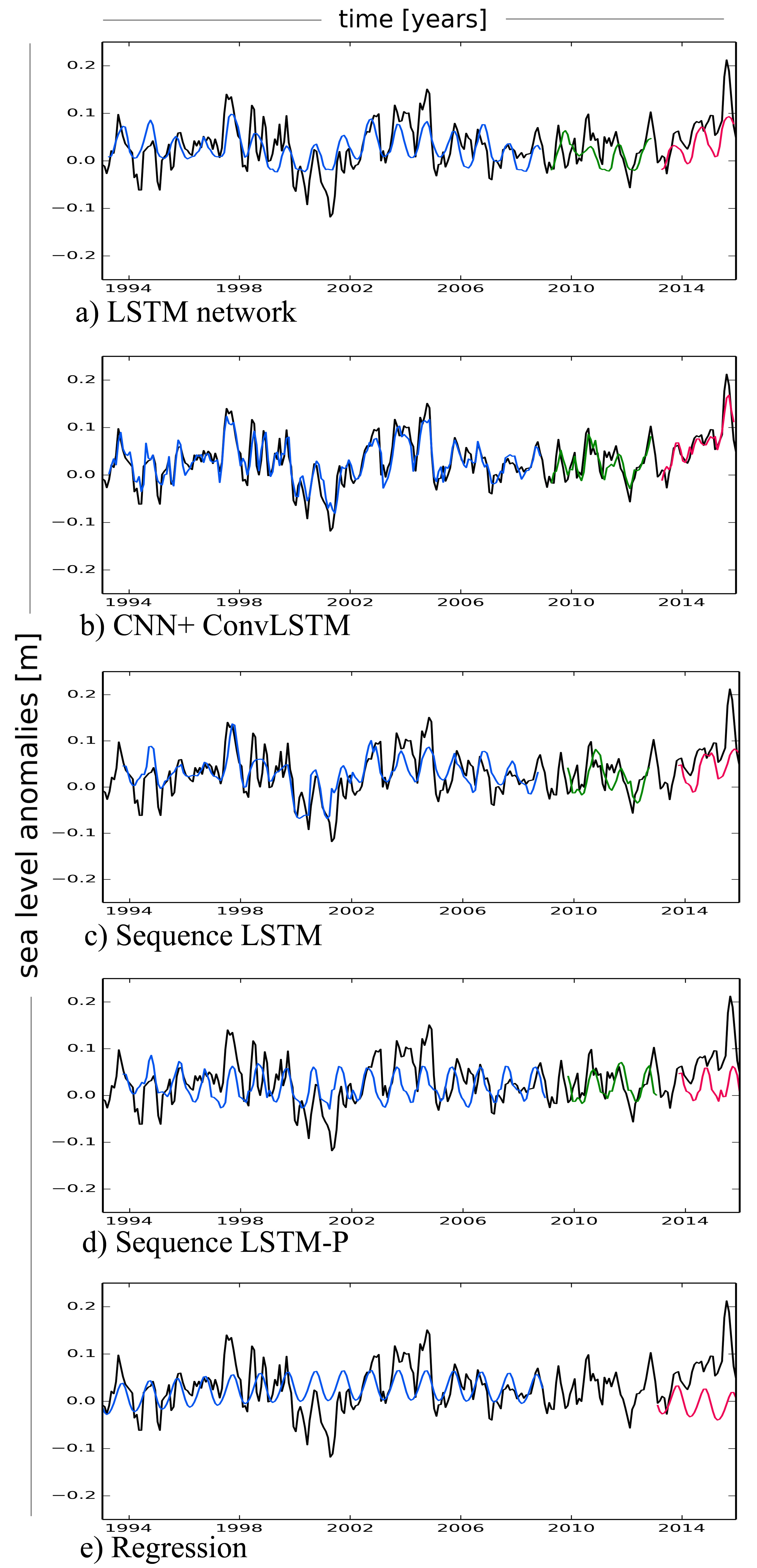}
	\vspace{-0.2cm}
	\captionof{figure}{Results for one grid cell (marked by a cross in Fig.~3): True (black) and predicted (blue = training, green = validation and red = test data) time series of SLAs}
%\label{fig:result}
\end{center}
\end{minipage}

\setlength{\belowrulesep}{0pt}
\hspace{-0.6cm}
\begin{table}
%\begin{small}
{\fontsize{8.5}{10} \selectfont 
\begin{tabular}{|c|c|c|c|}
\hline 
RMSE averaged over & training data & validation data & test data \\
\toprule[1pt]
LSTM network & 0.062~m & 0.071~m & 0.076~m \\ 
\hline 
CNN+ConvLSTM & 0.047~m & 0.050~m & 0.051~m \\ 
\hline 
Sequence LSTM & 0.059~m & 0.079~m & 0.077~m \\ 
\hline 
Sequence LSTM-P & 0.083~m & 0.080~m & 0.081~m \\ 
\hline 
Regression & 0.078~m & - & 0.154~m \\ 
\hline 
\end{tabular} 
%\end{small} 
}
\vspace{-0.35cm}
\caption{RMSEs between the networks' or the regression's prediction and the true SLAs averaged over all grid cells}
\end{table}

%\vspace{-0.8cm}

Table~1 shows the RMSEs between true SLAs and the predictions made by our networks or the regression. The training error of the \mbox{Sequence LSTM-P} is higher, because we here feed the predictions 20 times back into the network.

To examine the spatial patterns, we plot the true and predicted SLAs in November 2014 in Fig.~3. All networks are trained using data up to 2008. Starting the test phase in Ja-nuary 2013, the Sequence LSTM-P needs nine true months as input only once (i.e. till September 2013). The Sequence LSTM depends on true data every nine months, so June 2014 is the last true month needed to predict the sequence including November 2014. The LSTM and CNN+ConvLSTM depend on true inputs of the previous month (here October 2014).

Striking, overall all network architectures outperform the regression. Especially in Fig. 3f it can be seen that the regression fails at capturing the spatial structure. We observe that an additional CNN improves the accuracy significantly (compare Fig. 2a/b and 3b/c). In Fig. 3c the CNN+ConvLSTM network resolves nearly all spatial structures very well. However, we observe that in combination with a normal LSTM, the CNN brings no improvement (not shown here) - only if we also keep the spatial structure throughout the ConvLSTM. 

Both extending the prediction length to nine months (compare Fig. 2a/c and 3b/d)  and using the predicted values for the next prediction (compare Fig. 2c/d and 3d/e) leads to only slight degradation of predictive skill and the predictions stay close to the real values for a long time.

Figure 4 shows the spatial distribution of the RMSE \linebreak averaged over the test phase for the CNN+ConvLSTM and the Sequence LSTM-P. In both cases the main errors occur around the Kuroshio current with its highly variable eddies.

\section{Conclusion}
\label{sec:print}

In this work we develop a combined convolutional and recurrent neural network to analyse both the spatial and the temporal evolution of SLAs in the northern and central Pacific Ocean and to make accurate predictions for the future sea level. We validate the accuracy of our approach and compare the results to a regression. All the used network designs outperform the regression. Adding a convolutional neural network improves the accuracy significantly.
Therefore we plan to extend this architecture to take sequences as in- and output. We are confident that in this way we will be able to improve predictions at longer timescales. The developed method could also be applied to other regions or even globally.

\begin{figure*}
\centering
	\includegraphics[width=0.97\textwidth]{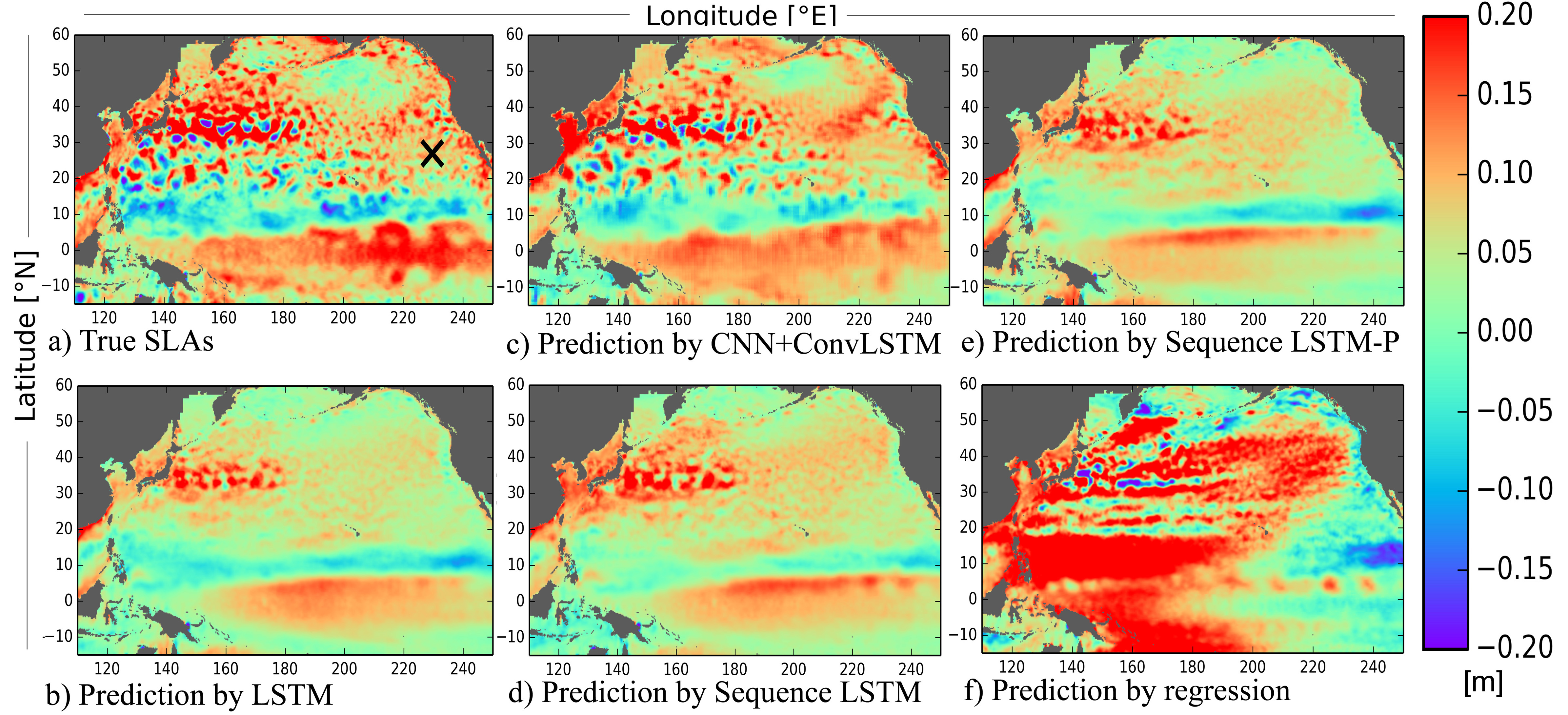}
\vspace{-0.2cm}
\caption{Spatial structure of the SLAs in the selected region in November 2014 (within the test period)}
\label{fig:result}
\end{figure*}

\begin{center}
\begin{minipage}[t!]{0.46\textwidth}
\centering
	\includegraphics[width=0.85\textwidth]{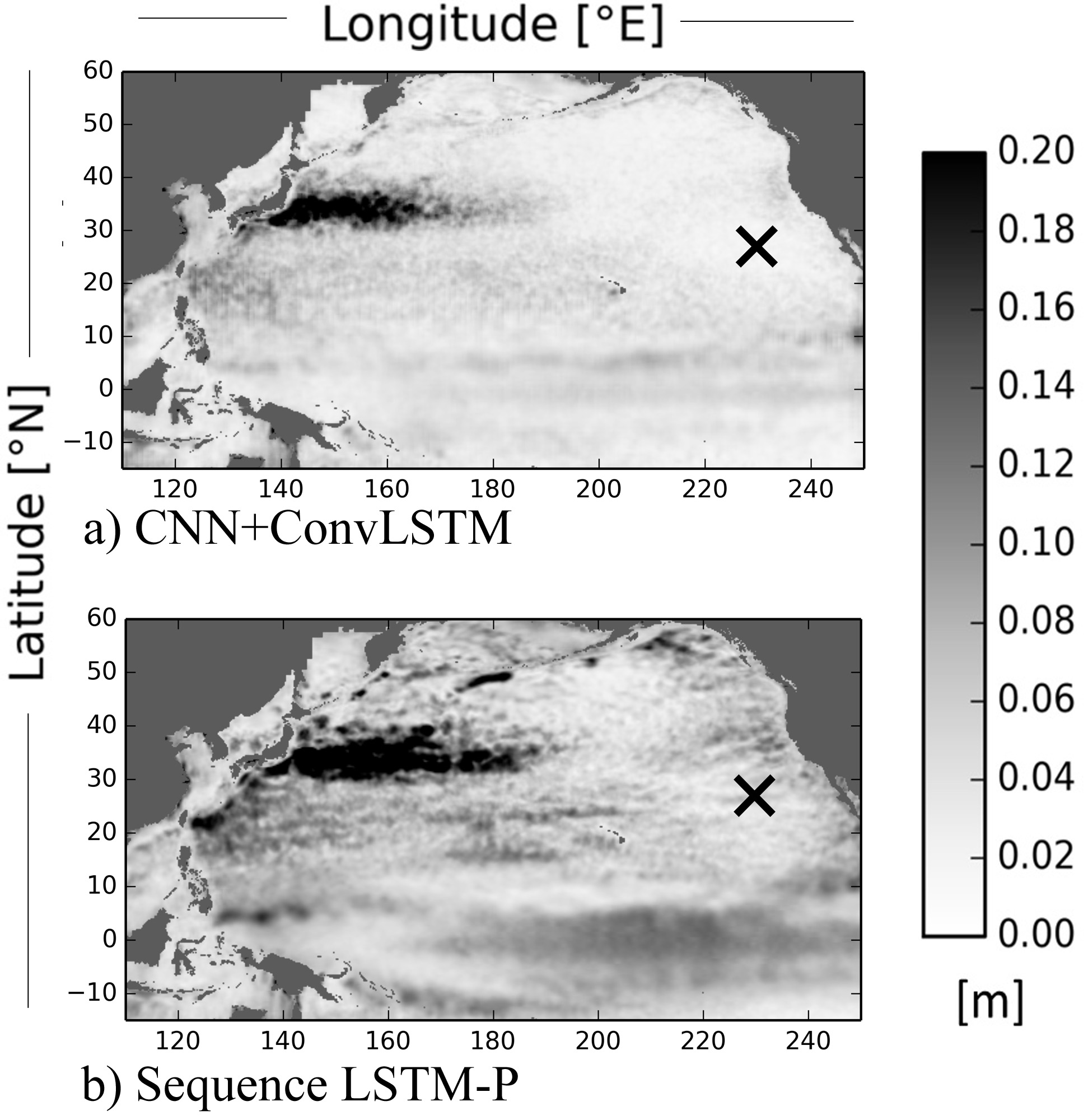}
	\vspace{-0.2cm}
	\captionof{figure}{RMSE at each grid cell averaged over the test phase}
\end{minipage}
\end{center}

\end{document}